\documentclass[lettersize,journal]{IEEEtran}

\usepackage{booktabs}
\usepackage{xcolor}
\usepackage{hyperref}

\ifCLASSINFOpdf
  \usepackage[pdftex]{graphicx}
  \graphicspath{{../figures/}}
\else
\fi
\begin{document}

\title{Do Satellite Tasks Need Special Pretraining?}

\author{Ani~Vanyan, 
        Alvard~Barseghyan, 
        Hakob~Tamazyan,
        Tigran~Galstyan, 
        Vahan~Huroyan, 
        Naira~Hovakimyan,
        Hrant~Khachatrian 
\thanks{A. Vanyan, A. Barseghyan, H. Tamazyan, H. Khachatrian are with the YerevaNN research lab and YSU.}
\thanks{V. Huroyan is with Saint Louis University.}
\thanks{N. Hovakimyan is with the University of Illinois at Urbana-Champaign.}
}


%



\maketitle

\begin{figure}
    \centering
    \includegraphics[clip, trim=2.5cm 19.50cm 6cm 2.75cm, width=0.5\textwidth]{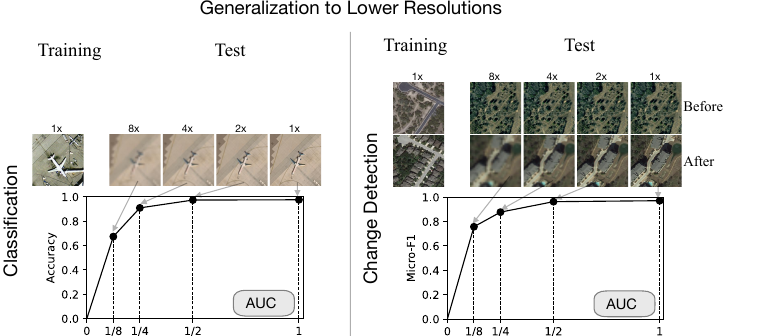}
    \label{fig:enter-label}
\end{figure}

\begin{abstract}
Foundation models have advanced machine learning across various modalities, including images. Recently multiple teams trained foundation models specialized for remote sensing applications. This line of research is motivated by the distinct characteristics of remote sensing imagery, specific applications and types of robustness useful for satellite image analysis.
In this work we systematically challenge the idea that specific foundation models are more useful than general-purpose vision foundation models, at least in the small scale. First, we design a simple benchmark that measures generalization of remote sensing models towards images with lower resolution for two downstream tasks. Second, we train iBOT, a self-supervised vision encoder, on MillionAID, an ImageNet-scale satellite imagery dataset, with several modifications specific to remote sensing. We show that none of those pretrained models bring consistent improvements upon general-purpose baselines at the ViT-B scale.
\end{abstract}

\begin{IEEEkeywords}
Remote sensing, vision transformers, self-supervised learning, change detection.
\end{IEEEkeywords}

%
\IEEEpeerreviewmaketitle

\section{Introduction}

The rapid advancements in remote sensing technologies have led to an increased reliance on foundation models for interpreting vast amounts of imagery data captured by remote sensing satellites.
Usually, this data is raw and unlabeled, whereas creating labels is time-consuming and expensive. 
Many critical tasks, like change detection, image classification, and semantic segmentation, applied for land cover mapping, disaster monitoring, urban growth, vegetation health, and terrain analysis, require labeled data for effective model training.
In line with recent advancements in self-supervised and semi-supervised learning for vision tasks, the current trend is to train a self-supervised model (either contrastive or based on masked image modeling) which later serves as a backbone for fine-tuning for subsequent downstream tasks.


Most of the publicly available satellite imagery comes from Sentinel-1 and Sentinel-2 which provide quite low resolution images. As these images are not detailed enough for many applications, even for human eyes, most of the expert annotations are collected on higher resolution images, which are rare and are not always available for deployed systems. This prompts a specific requirement for remote sensing foundation models: the fine-tuned versions should properly generalize to images with lower resolution than the ones labeled for fine-tuning. To evaluate this kind of generalization, we design a simple benchmark which covers scene classification~\cite{resisc,ucmerced} and change detection~\cite{LEVIR,CDD} tasks.

We take an off-the-shelf, general-purpose visual foundation model, iBOT~\cite{iBot}, which is trained on ImageNet containing little-to-no satellite imagery. Then we pretrain another iBOT on MillionAID~\cite{million_aid}, an ImageNet-scale dataset containing remote sensing images from various satellites, and widely used in specialized foundation models~\cite{cmid}. Additionally, we implement two modifications to the iBOT training. (a) we use image-scale augmentations during pretraining, to verify its effect on downstream generalization capabilities. (b) as many remote sensing tasks involve dense prediction, e.g. change detection, it requires relatively large heads to be trained from scratch. We create an artificial task to pretrain a head for change detection with purely unlabeled data.

Finally, we compare our models with a few publicly available remote sensing foundation models. The results indicate that there is no consistent benefit from pretraining on remote sensing data, as well as the additional tricks we suggested. On the other hand, general purpose foundation models keep strengthening over time, and it is increasingly harder for specialized models to be competitive . 

Note that we limit our analysis to small models, particularly to ViTs with less than 100M parameters. While we limit the FLOPs for processing a single image, we do not limit the amount of compute used for training them. Particularly, we compare to the ViT-B version of DINOv2~\cite{oquab2023dinov2}, which is distilled from a larger ViT-g model. While the large model was trained using hundreds of GPUs, the distilled version can be easily fine-tuned on a single consumer-grade GPU.


\section{Related Work}

Some recent developments in the field include various approaches using either supervised or self-supervised learning algorithms.
Surprisingly, for some transformer-based models, performance on ImageNet in certain instances outperforms those pre-trained on remote sensing imagery~\cite{vanyan2023analyzing}. 
The effect of pre-training on ImageNet  vs a large remote sensing scene recognition dataset is studied in~\cite{rsp}.
To serve as a pre-training dataset, some existing techniques involve gathering data from available open-source large remote sensing datasets and employing it to train the self-supervised algorithm. 
The two main methods to train self-supervised foundation models are contrastive learning-based methods and generative-based methods (masked image modeling).

Similar to classical contrastive learning-based methods, recent advancements include SECO~\cite{SECO}, CACo~\cite{caco}, MATTER~\cite{matter}, Dino-MC~\cite{dino-mc}, among others.
Another line of research builds on Masked Autoencoders (MAE)~\cite{MAE}, a successful foundation model utilizing masked image modeling, where the pretext task is to reconstruct an image from its masked version. Notable extensions include SatMAE~\cite{SatMAE}, Scale-MAE~\cite{scalemae}, 
and SpectralGPT~\cite{spectralgpt}.
A more recent direction aims to integrate reconstruction-based and contrastive learning-based approaches. Notable examples include CMID~\cite{cmid}, GFM~\cite{gfm}, SECO~\cite{SECO}, and CROMA~\cite{croma}. 
\cite{gfm} observed that some state-of-the-art methods for aerial imagery often do not outperform ImageNet-22k pretrained ViTs.
Another research focus is multi-task pretraining, with works such as Satlas~\cite{SatlasPretrain} and MTP~\cite{mtp}. 
Recently, for change detection, an end-to-end super-resolution-based network, SRCDNet~\cite{srcd}, was introduced to address change detection across varying image resolutions. We extend this idea to additional classification and change detection datasets.


\section{The Benchmark}
\label{sec:benchmark}

Generalization can be evaluated across various aspects, including adaptation to different spatial resolutions, spectral bands, seasonal variations, times of day, and diverse geographical locations. 
In this work, we focus on evaluating the foundation model's ability to generalize to unseen resolutions across two key tasks: scene classification and change detection.
We emphasize that our evaluation focuses solely on \textbf{generalization to lower spatial resolutions}.
Low-resolution satellites, such as Landsat and Sentinel, provide publicly available imagery, whereas higher-resolution imagery is often more difficult to obtain. In many scenarios, image labeling is performed on high-quality imagery, as it is often hard to see necessary details on low resolution images even for human annotators. But at test time, the images may come from satellites with lower resolution. Therefore, we expect models to perform robustly under such distribution shifts.
While generalization to higher spatial resolutions can also occur in practical applications, retaining performance at higher resolutions is trivial by simply downsampling images to the original resolution.

\noindent\textbf{Datasets.} 
RESISC45 \cite{resisc} and UC Merced \cite{ucmerced} datasets contain 256x256px images. Image resolution is 30cm/px for UC Merced and varies 20-600cm/px for RESISC45. Both datasets use RGB bands only. We take the splits defined in \cite{in-domain-repr-learning-2019}. \\
The LEVIR-CD dataset~\cite{LEVIR} comprises a substantial collection of bitemporal Google Earth images. It includes $637$ image pairs, each sized $1024\times1024$px, with $400$ images designated for training. 
The images in the training set have a resolution of 50cm/px. 
The fully annotated LEVIR-CD dataset encompasses a total of $31,333$ individual changed buildings. 
The changes in the LEVIR-CD dataset primarily come from the construction of new buildings. 
The average size of each changed area is approximately $987$ pixels.
\\
The CDD~\cite{CDD} dataset contains season-varying remote sensing images of the same region, obtained from Google Earth (DigitalGlobe). The dataset comprises $16,000$ image sets (two images of the same location and the annotated change), each with an image size of $256 \times 256$ pixels and 0.03-1m/px ground sample distance.

\noindent\textbf{Scene Classification.}
We use two commonly used benchmark datasets in the literature: RESISC45~\cite{resisc} and UC Merced~\cite{ucmerced}; see Sec.~\ref{sec:benchmark}.
Performance is measured at the original resolution and at reduced resolutions
(1/2, 1/4 and 1/8). 
Images are downscaled by a factor of $1/x$ and then upscaled back by  $x$, preserving pixel count but reducing quality. 
This simulates lower-resolution satellite imagery.
As an evaluation metric, we plot a curve with the scaling factor ($1/8$, $1/4$, $1/2$, $1$) on the x-axis and accuracy on the y-axis. The area under this curve (\textbf{AUC-Acc}) serves as our final metric.
We restrict the models to use 50 GFLOPs on a single image. This threshold is independent from the neural architecture, and ViT-B/16 on an image of size 256x256px is within the limits.

\noindent\textbf{Change Detection.}
We use two commonly used datasets: CDD~\cite{CDD} and LEVIR-CD~\cite{LEVIR}; see Sec.~\ref{sec:benchmark}. 
We create partially scaled versions of the test sets of these datasets. We maintain the scale of the first image unchanged, while for the second image, we distort it by reducing its quality by a factor of 2, 4, and 8. Note that a similar setup has been first proposed in \cite{srcd}. We evaluate on the original resolution, as well as on the scaled versions. We compute micro-averaged F1 score for each of the versions. Finally we draw a curve where x-axis is the scaling parameter and y-axis is the micro-averaged F1 score for each version. We report the area under this curve as our final metric, and call it \textbf{AUC-F1}.
For this benchmark, we restrict the models to use 100 GFLOPs on a pair of images.

\section{Pretraining iBOT on Remote Sensing Data}
\label{sec:contributing_factors}

\noindent\textbf{iBOT pretraining.} 
~\cite{vanyan2023analyzing} showed that self-distillation models generally outperform MIM-based models in learning robust image representations especially at the level of patch representations. We chose a typical self-distillation method with a publicly available codebase, iBOT, as a basis for our experiments.
We pre-trained iBOT with the MillionAID dataset~\cite{million_aid}, dividing images into a maximum of 550-pixel square tiles, yielding $2 106 700$ images. 
We trained iBOT for 200 epochs with peak learning rate $5 \times 10^{-4}$ that linearly decreases to $2 \times 10^{-6}$ over 5 warmup epochs. All RandomResizeCrops were converted to RandomCrops in the transforms. The training was conducted using PyTorch Distributed Data Parallel to utilize multiple GPUs and used 100 batch size per GPU. The experiments were performed on NVIDIA DGX A100 at the local university 
and an instance with 8 NVIDIA H100s kindly provided by Nebius.ai. 
The resulting model is labeled as \textbf{iBOT-MillionAID}. The original iBOT pre-trained on ImageNet is served as a baseline.



\noindent\textbf{Augmentation.} 
We analyze scale augmentation's impact on robustness to scale changes. iBOT's augmentation module resizes and crops images. 
We pre-trained two iBOTs: with and without resizing.
The hypothesis is that scale augmentation improves robustness, transferring to fine-tuned models and increasing AUC scores on our benchmark.
We also test scale augmentation during fine-tuning by shrinking images (or the second image in change detection) by 2, 4, and 8 times, then resizing them back.

Table~\ref{tab:augm} shows that scale augmentation during pretraining still does not improve generalization capabilities, while augmentation during fine-tuning consistently and significantly improves the scores of our benchmark.

\begin{table}[h!]
\centering
\caption{Dependence of the performance of fine-tuned models on scale augmentation performed during pretraining and fine-tuning. All models are iBOTs trained on MillionAID.}
\label{tab:augm}
\resizebox{0.5\textwidth}{!}{%
\begin{tabular}{lcccc|c}
\toprule
\multicolumn{1}{c}{\underline{\textbf{Augmentation Phase}}} & 1:1 & 1:2 & 1:4 & 1:8 &  \\\midrule
\textbf{LEVIR-CD}&  &  &  &  & AUC-F1 \\
\midrule
{\color{gray} Pretraining} / {\color{gray} Fine-tuning} & $88.7 \pm 0.1$ & $86.5 \pm 0.2$ & $63.6 \pm 3.3$ & $7.5 \pm 0.5$ & $67.5 \pm 0.7$ \\
\underline{\textbf{Pretraining}} / {\color{gray} Fine-tuning} & $90.6 \pm 0.2$ & $87.6 \pm 0.9$ & $50.4 \pm 15.1$ & $2.0 \pm 1.0$ & $65.2 \pm 3.2$ \\
{\color{gray} Pretraining} / \underline{\textbf{Fine-tuning}} & $88.2 \pm 0.1$ & $88.4 \pm 0.1$ & $87.9 \pm 0.1$ & $86.1 \pm 0.1$ & $82.4 \pm 0.1$ \\
\underline{\textbf{Pretraining}} / \underline{\textbf{Fine-tuning}} & $89.9 \pm 0.1$ & $89.9 \pm 0.1$ & $89.4 \pm 0.1$ & $87.7 \pm 0.1$ & $83.9 \pm 0.1$ \\
\midrule
\textbf{UC Merced} &   &  &  &  & AUC-ACC \\
\midrule
{\color{gray} Pretraining} / {\color{gray} Fine-tuning} & $98.0 \pm 0.3$ & $97.2 \pm 0.6$ & $87.2 \pm 1.9$ & $38.7 \pm 3.0$ & $82.2 \pm 0.7$ \\
\underline{\textbf{Pretraining}} / {\color{gray} Fine-tuning} & $98.7 \pm 0.8$ & $97.9 \pm 1.3$ & $84.3 \pm 4.3$ & $46.0 \pm 8.3$ & $82.9 \pm 1.0$ \\
{\color{gray} Pretraining} / \underline{\textbf{Fine-tuning}}  & $98.2 \pm 0.6$ & $98.3 \pm 0.6$ & $98.0 \pm 0.6$ & $95.7 \pm 1.2$ & $91.8 \pm 0.6$ \\
\underline{\textbf{Pretraining}} / \underline{\textbf{Fine-tuning}} & $95.3 \pm 1.8$ & $94.7 \pm 2.0$ & $94.0 \pm 2.4$ & $91.8 \pm 3.6$ & $88.4 \pm 2.1$ \\
\bottomrule
\end{tabular}
}

\end{table}

\begin{table}[b]
    \centering
    
    \caption{The effect of a pretrained mask decoder on change detection tasks. All models are iBOTs pretrained on MillionAID with scale augmentation.}
    \label{tab:decoder}
    \resizebox{0.5\textwidth}{!}{%
    \begin{tabular}{lccccc}
        \toprule
        \textbf{LEVIR-CD} & 1:1 & 1:2 & 1:4 & 1:8 & AUC-F1 \\
        \midrule
        Without Mask Decoder & $90.6 \pm 0.2$ & $87.6 \pm 0.9$ & $50.4 \pm 15.1$ & $2.0 \pm 1.0$ & $65.2 \pm 3.2$ \\
        With Mask Decoder & $90.6 \pm 0.1$ & $89.2 \pm 0.1$ & $66.6 \pm 5.0$ & $4.3 \pm 1.1$ & $69.1 \pm 1.0$ \\
        \midrule
        \textbf{CDD} & \multicolumn{4}{c}{} &  \\
        \midrule
        Without Mask Decoder & $97.4 \pm 0.0$ & $96.8 \pm 0.0$& $91.4 \pm 0.6$ & $79.2 \pm 0.9$ & $87.7 \pm 0.2$ \\
        With Mask Decoder & $97.1 \pm 0.0$ & $96.7 \pm 0.0$ & $91.5 \pm 0.5$ & $80.1 \pm 0.9$ & $87.7 \pm 0.2$ \\
        \bottomrule
    \end{tabular}
    }
\end{table}

\begin{table}[ht]
    \centering
    \caption{The impact of full fine-tuning.
    All models are iBOTs pretrained on MillionAID with scale augmentation. No scale-augmentation was performed after pretraining.}
    \label{tab:frozen}
    \resizebox{0.5\textwidth}{!}{%
    \begin{tabular}{lccccc}
        \toprule
        \textbf{RESISC45} & \multicolumn{4}{c}{} & AUC-ACC \\
        \midrule
        Full fine-tuning & $93.4 \pm 0.2$ & $84.3 \pm 1.2$ & $47.4 \pm 5.6$ & $18.7 \pm 2.0$ & $66.2 \pm 1.8$  \\
        Frozen backbone & $\mathbf{94.6 \pm 0.1}$ & $\mathbf{92.2 \pm 0.2}$ & $\mathbf{66.5 \pm 1.5}$ & $\mathbf{25.1 \pm 1.3}$ & $\mathbf{73.8 \pm 0.5}$ \\
        \midrule
        \textbf{LEVIR-CD} & 1:1 & 1:2 & 1:4 & 1:8 & AUC-F1 \\
        \midrule
        Full fine-tuning & $90.6 \pm 0.2$ & $87.6 \pm 0.9$ & $50.4 \pm 15.1$ & $2.0 \pm 1.0$ & $65.2 \pm 3.2$ \\
        Frozen backbone & $84.4 \pm 0.0$ & $84.4 \pm 0.2$ & $61.6 \pm 7.8$ & $3.4 \pm 4.0$ & $64.7 \pm 2.0$ \\
        \midrule
        \textbf{UC Merced} & \multicolumn{4}{c}{} & AUC-ACC \\
        \midrule
        Full fine-tuning & $98.7 \pm 0.8$ & $97.9 \pm 1.3$ & $84.3 \pm 4.3$ & $46.0 \pm 8.3$ & $82.9 \pm 1.0$ \\
        Frozen backbone & $99.5 \pm 0.1$ & $99.2 \pm 0.3$ & $75.7 \pm 2.9$ & $31.3 \pm 3.9$ & $80.2 \pm 0.7$ \\
        \bottomrule
    \end{tabular}
    }

\end{table}

\begin{figure}
    \centering
    \includegraphics[clip, trim=1.7cm 19.5cm 3cm 3.5cm, width=0.5\textwidth]{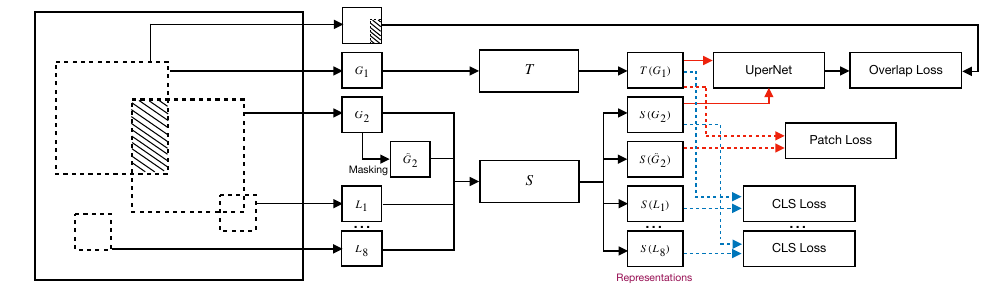}
    \caption{iBOT pretraining architecture with an additional UperNet mask decoder that is trained using the ``overlap loss''. There are two global and eight local crops of the original image that pass through Teacher (T) and Student (S) networks. 
    Dotted lines imply that only the representations of the last layers are used. Solid lines imply that representations of four layers are used (as an input to UperNet). Red lines correspond to patch representations, the blue lines correspond to CLS vectors.}
    \label{fig:model-architecture}
\end{figure}

\noindent\textbf{Pretrained mask decoder.} 
We extend iBOT-MillionAID with a pretrained mask decoder for segmentation and change detection tasks, requiring a binary mask, and leverage a module pretrained on large datasets. The teacher processes two global crops, while the student handles those plus eight local crops.
Since MillionAID lacks segmentation masks, we map the second global crop’s mask to the first crop’s coordinate space as the target mask. Patch representations from both crops are concatenated and fed into an UperNet~\cite{upernet}  decoder to generate the binary mask with a pixel-wise cross-entropy loss. 
The architecture and details are in Fig.~\ref{fig:model-architecture} and Sec.~\ref{sec:implementation}.

As shown in Table~\ref{tab:decoder}, there is a slight improvement in performance and significantly lower variance across all scales with the pretrained mask decoder on LEVIR-CD. There is no visible change on CDD. This can be explained by the large size of the CDD dataset.
It is likely that the additional power of the pretrained models is not critical when the fine-tuning dataset is large enough.
Another way to enhance the impact of pretrained decoders is to pretrain it with denser supervision signal. While we used a binary mask calculated during pretraining, \cite{mtp} uses segmentation pseudo-labels generated by a strong domain-agnostic segmentation model.

\noindent\textbf{Catastrophic Forgetting During Fine-Tuning.}
Pretrained models may lose generalization during fine-tuning. To assess this, we repeat fine-tuning with frozen backbones, ensuring the final linear layer or decoder lacks exposure to diverse scales.
Table~\ref{tab:frozen} shows that the effect varies by dataset. 
For RESISC45, freezing the backbone improves robustness to lower resolutions. 
LEVIR-CD follows this trend at 1:4 and 1:8 resolutions, though full fine-tuning performs better at 1:1 and 1:2. 
In contrast, UC Merced benefits from a frozen backbone at higher resolutions, while full fine-tuning excels at lower resolutions.

\noindent\textbf{More methods to compare.}
We compared our pretrained iBOT with SatlasPretrain \cite{SatlasPretrain} trained on high-resolution imagery (Aerial) and on the RGB subset of Sentinel-2 imagery (S2), GFM \cite{gfm}, and general-purpose iBOT pretrained on ImageNet.
Each of these models have a different training paradigm and pretraining dataset. iBot is a self-supervised method pretrained on ImageNet. 
GFM combines two concepts: self-supervised pretraining on a custom-collected dataset, GeoPile, and continual pretraining to retain knowledge obtained from pretraining on ImageNet. SatlasPretrain is pretrained on a custom-collected dataset, Satlas, in a supervised manner. 
Clay v1~\cite{clay} is a self-supervised method that utilizes a hybrid loss combining distillation and reconstruction components. 
Prithvi {\cite{prithvi}} is a modification of a MAE model to support 3D inputs with 6 channels.
We adapt all these models to work with the datasets used in our benchmark. 

\begin{figure*}
  \centering
  \includegraphics[width=0.22\linewidth]{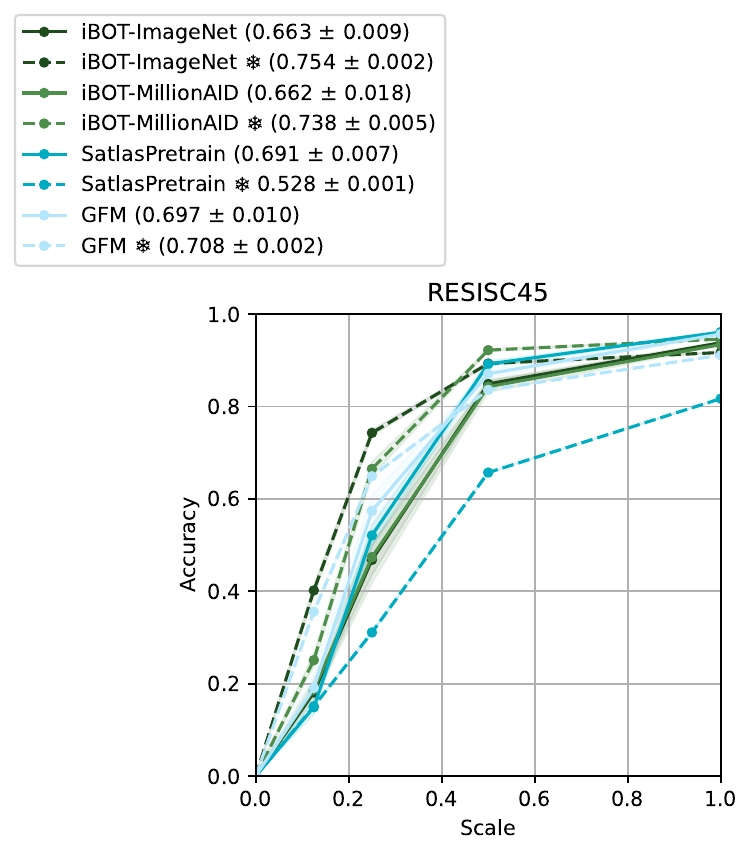}
  \includegraphics[width=0.22\linewidth]{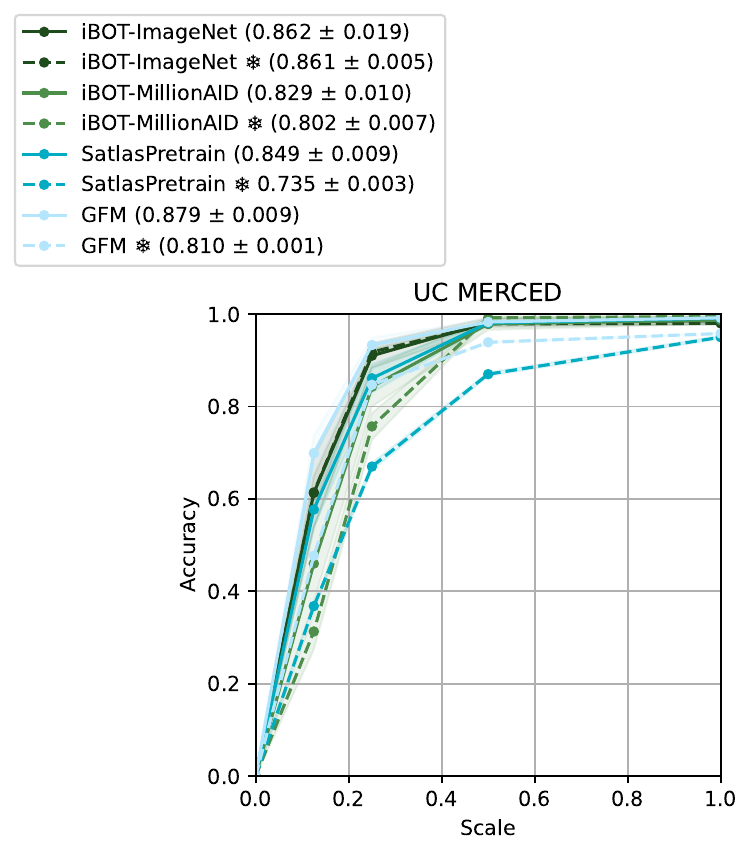}
  \includegraphics[width=0.22\linewidth]{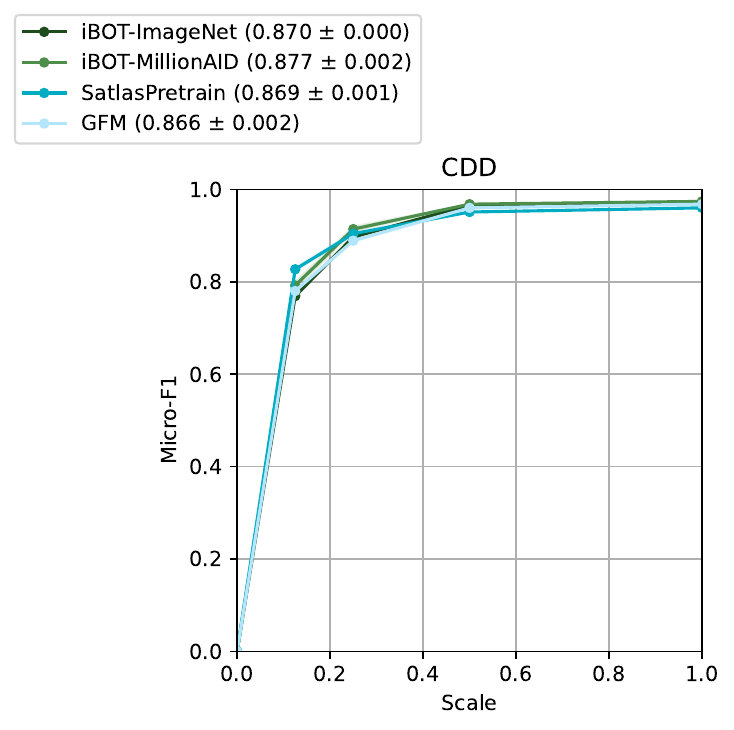}
  \includegraphics[width=0.22\linewidth]{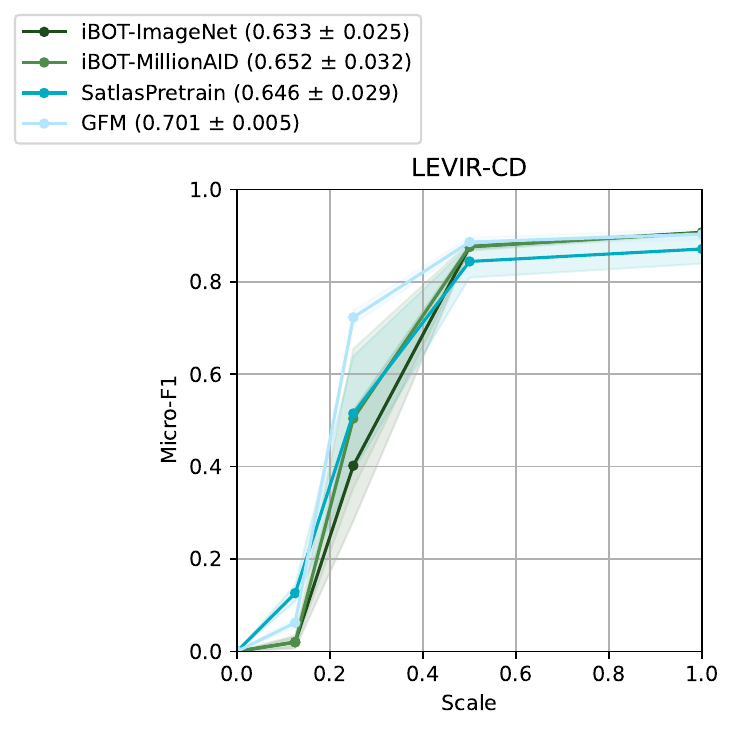}
  \caption{The results of the baselines on our benchmark tasks for generalization across image resolution. The top row shows classification on RESISC and UC Merced, while the bottom row shows change detection on CDD~ and LEVIR-CD. 
  X-axis: Scale of Distortions, Y-axis: Micro-F1 Scores.}
  \label{fig:results-auc}
\end{figure*}





\section{Implementation Details}
\label{sec:implementation}

To adapt the models for classification, we add a linear layer on top of the [CLS] token representation, if available, or on top of the global average pooled vector of all patch representations.
To test the models for change detection, we take the backbone, which is either a Swin Transformer, or a ViT, and integrate the UperNet  head~\cite{upernet}. The two source images go through identical backbones, and the resulting representations are substracted from each other and passed to the head.
In the case of ViTs, we use an additional \textit{neck} module between the backbone and UperNet. The backbone is initialized with the pre-trained weights and further fine-tuned using the change detection datasets. In case of our iBOT trained on MillionAID, the neck and the head modules are also initialized, and we take the concatenation of features instead of the difference. 
All the codes for pretraining, as well as the benchmarks proposed by us with all the hyperparameters, can be found at: \url{https://github.com/YerevaNN/rs_foundation_models}.

\noindent\textbf{Classification:} We perform two kinds of fine-tuning: full fine-tuning and linear probing. For both setups, we train for 100 epochs. For all experiments in the full fine-tuning setup or linear probing, we evaluate using the last checkpoint (except for full fine-tuning on the BigEarthNet dataset, where we select the best checkpoint based on the validation set performance). In all experiments within the full fine-tuning setup, we use the AdamW optimizer with a learning rate of $10^{-4}$ employing Warmup Cosine scheduler and an estimated minimum value of $10^{-5}$. In experiments within the linear probing setup, we use the AdamW optimizer with a learning rate of $10^{-3}$ employing MultiStep scheduling and an estimated minimum value of $10^{-5}$. For Prithvi and Channel-ViT we did an extra tuning of hyperparameters, and switched the scheduler to Warmup Cosine for Prithvi, and switched to Adam for Channel-ViT.


\noindent\textbf{Change Detection:}
For change detection experiments, we train our models for $200$ epochs.
We use the AdamW optimizer with a Warmup Cosine scheduler (peak learning rate: $6 \times 10^{-5}$) which includes warmup steps of $10$ and batch size of $32$. 

\noindent\textbf{Pretrained Mask Decoder}
Note that UperNet uses features from ViT layers 3, 5, 8, and 12.
We explored two methods to integrate mask loss into iBOT training: using only the student for patch representations or incorporating the teacher for one. The first approach led to unstable training with spiking activations, while the teacher-student method ensured stable joint training.
We used $2.5 \times 10^{-4}$ peak learning rate and cosine decay with 5 warmup epochs.

\section{Results}

\begin{table}[ht]
    \centering
    \caption{Benchmark Results for Change Detection (LEVIR-CD, CDD) and Classification (RESISC45, UC Merced) tasks with Different Scale Distortions.}
    \label{tab:results-full}
    \resizebox{0.5\textwidth}{!}{%
    \begin{tabular}{lccccc}
        \toprule
        \textbf{LEVIR-CD} & \textbf{1:1} & \textbf{1:2} & \textbf{1:4} & \textbf{1:8} & \textbf{AUC-F1} \\
        \midrule
        iBOT-ImageNet & $90.7 \pm 0.1$ & $87.6 \pm 0.5$ & $40.2 \pm 12.0$ & $2.0 \pm 1.4$ & $63.3 \pm 2.5$ \\
        iBOT-MillionAID & $90.6 \pm 0.2$ & $87.6 \pm 0.9$ & $50.4 \pm 15.1$ & $2.0 \pm 1.0$ & $65.2 \pm 3.2$ \\
        SatlasPretrain (S2\_SwinB\_SI\_RGB) & $87.1 \pm 3.2$ & $84.4 \pm 3.5$ & $51.5 \pm 12.4$ & $12.6 \pm 1.8$ & $64.6 \pm 2.9$ \\
        GFM & $90.3 \pm 1.1$ & $88.6 \pm 1.0$ & $72.3 \pm 1.5$ & $6.2 \pm 1.1$ & $70.1 \pm 0.5$ \\
        Prithvi & $85.2 \pm 0.1$ & $84.4 \pm 0.1$ & $76.4 \pm 1.1$ & $14.5 \pm 1.2$ & $69.1 \pm 0.4$ \\
        {DINOv2} & $88.0 \pm 0.1$ & $86.5 \pm 0.2$ & $70.4 \pm 1.5$ & $12.2 \pm 2.5$ & $69.1 \pm 0.6$ \\
        \midrule
        \textbf{CDD} & \multicolumn{4}{c}{} & \textbf{AUC-F1} \\
        \midrule
        iBOT-ImageNet & $97.3 \pm 0.0$ & $96.6 \pm 0.0$ & $89.7 \pm 0.2$ & $76.9 \pm 0.4$ & $87.0 \pm 0.0$ \\
        iBOT-MillionAID & $97.4 \pm 0.0$ & $96.8 \pm 0.0$& $91.4 \pm 0.6$ & $79.2 \pm 0.9$ & $87.7 \pm 0.2$ \\
        SatlasPretrain (S2\_SwinB\_SI\_RGB) & $96.0 \pm 0.0$ & $95.1 \pm 0.0$ & $90.4 \pm 0.3$ & $82.7 \pm 0.4$ & $86.9 \pm 0.1$ \\
        GFM & $96.8 \pm 0.0$ & $96.0 \pm 0.1$ & $88.9 \pm 0.3$ & $78.0 \pm 0.6$ & $86.6 \pm 0.2$ \\
        Prithvi & $90.9 \pm 0.2$ & $90.5 \pm 0.2$ & $88.5 \pm 0.3$ & $82.9 \pm 0.8$ & $83.6 \pm 0.3$ \\
        {DINOv2} & $92.4 \pm 0.0$ & $91.3 \pm 0.1$ & $87.5 \pm 0.1$ & $78.2 \pm 0.1$ & $83.5 \pm 0.0$ \\
        \midrule
        \textbf{RESISC45: full fine-tuning} & \multicolumn{4}{c}{} & \textbf{AUC-ACC} \\
        \midrule
        iBOT-ImageNet & $93.8 \pm 0.2$ & $84.9 \pm 0.8$ & $46.8 \pm 3.3$ & $18.1 \pm 0.7$ & $66.3 \pm 0.9$ \\
        iBOT-MillionAID & $93.4 \pm 0.2$ & $84.3 \pm 1.2$ & $47.4 \pm 5.6$ & $18.7 \pm 2.0$ & $66.2 \pm 1.8$ \\
        DINOv2 & $94.1 \pm 0.4$ & $84.3 \pm 1.7$ & $46.7 \pm 5.2$ & $19.3 \pm 2.6$ & $66.3 \pm 1.6$ \\
        SatlasPretrain (S2\_SwinB\_SI\_RGB) & $96.1 \pm 0.1$ & $89.2 \pm 1.2$ & $61.4 \pm 3.3$ & $23.7 \pm 2.6$ & $71.9 \pm 1.4$ \\
        SatlasPretrain (Aerial\_SwinB\_SI) & $96.1 \pm 0.1$ & $89.2 \pm 0.6$ & $52.1 \pm 2.3$ & $14.9 \pm 1.5$ & $69.1 \pm 0.7$ \\
        GFM & $95.7 \pm 0.1$ & $87.1 \pm 0.9$ & $57.4 \pm 3.4$ & $19.1 \pm 3.0$ & $69.7 \pm 1.0$ \\
        \midrule
        \textbf{RESISC45: linear probing} & \multicolumn{4}{c}{} & \textbf{AUC-ACC} \\
        \midrule
        iBOT-ImageNet & $91.7 \pm 0.1$ & $89.3 \pm 0.2$ & $74.3 \pm 0.6$ & $40.2 \pm 0.9$ & $75.4 \pm 0.2$ \\
        iBOT-MillionAID & $94.6 \pm 0.1$ & $92.2 \pm 0.2$ & $66.5 \pm 1.5$ & $25.1 \pm 1.3$ & $73.8 \pm 0.5$ \\
        DINOv2 & $91.1 \pm 0.7$ & $87.2 \pm 1.0$ & $72.9 \pm 1.4$ & $40.3 \pm 1.0$ & $74.2 \pm 0.9$ \\
        SatlasPretrain (S2\_SwinB\_SI\_RGB) & $72.8 \pm 0.1$ & $58.0 \pm 0.2$ & $25.4 \pm 0.4$ & $15.0 \pm 0.3$ & $46.6 \pm 0.1$ \\
        SatlasPretrain (Aerial\_SwinB\_SI) & $81.7 \pm 0.1$ & $65.7 \pm 0.1$ & $31.1 \pm 0.3$ & $15.1 \pm 0.1$ & $52.8 \pm 0.1$ \\
        GFM & $91.1 \pm 0.0$ & $83.6 \pm 0.1$ & $64.9 \pm 0.4$ & $35.6 \pm 0.6$ & $70.8 \pm 0.2$ \\
        \midrule
        \textbf{UC Merced: full fine-tuning} & \multicolumn{4}{c}{} & \textbf{AUC-ACC} \\
        \midrule
        iBOT-ImageNet & $98.6 \pm 0.7$ & $98.2 \pm 1.0$ & $91.0 \pm 2.7$ & $61.3 \pm 7.7$ & $86.2 \pm 1.9$ \\
        iBOT-MillionAID & $98.7 \pm 0.8$ & $97.9 \pm 1.3$ & $84.3 \pm 4.3$ & $46.0 \pm 8.3$ & $82.9 \pm 1.0$ \\
        DINOv2 & $98.1 \pm 0.5$ & $97.9 \pm 0.3$ & $98.1 \pm 0.4$ & $97.3 \pm 0.3$ & $91.8 \pm 0.1$ \\
        SatlasPretrain (S2\_SwinB\_SI\_RGB) & $98.7 \pm 0.2$ & $98.0 \pm 0.3$ & $87.3 \pm 2.6$ & $61.9 \pm 5.9$& $85.5 \pm 1.3$ \\
        SatlasPretrain (Aerial\_SwinB\_SI) & $99.1 \pm 0.2$ & $98.1 \pm 0.3$ & $86.1 \pm 3.1$ & $57.7 \pm 3.9$ & $84.9 \pm 0.9$ \\
        GFM & $99.2 \pm 0.2$ & $98.3 \pm 0.6$ & $93.3 \pm 1.6$ & $69.9 \pm 3.8$ & $87.9 \pm 0.9$ \\
        \midrule
        \textbf{UC Merced: linear probing} & \multicolumn{4}{c}{} & \textbf{AUC-ACC} \\
        \midrule
        iBOT-ImageNet & $98.0 \pm 0.3$ & $97.9 \pm 0.3$ & $91.8 \pm 0.7$ & $61.4 \pm 3.6$ & $86.1 \pm 0.5$ \\
        iBOT-MillionAID & $99.5 \pm 0.1$ & $99.2 \pm 0.3$2 & $75.7 \pm 2.9$ & $31.3 \pm 3.9$ & $80.2 \pm 0.7$ \\
        DINOv2 & $97.4 \pm 0.2$ & $97.0 \pm 0.1$ & $96.8 \pm 0.1$ & $91.8 \pm 0.4$ & $90.3 \pm 0.1$ \\
        SatlasPretrain (S2\_SwinB\_SI\_RGB) & $85.7 \pm 0.8$ & $79.6 \pm 0.4$ & $55.6 \pm 1.6$ & $27.2 \pm 0.5$ & $65.1 \pm 0.3$ \\
        SatlasPretrain (Aerial\_SwinB\_SI) & $95.0 \pm 0.3$ & $87.0 \pm 0.4$ & $67.0 \pm 0.8$ & $36.8 \pm 0.3$ & $73.5 \pm 0.3$ \\
        GFM & $95.8 \pm 0.1$ & $93.9 \pm 0.2$ & $84.7 \pm 0.4$ & $47.7 \pm 0.4$ & $81.0 \pm 0.1$ \\
        \bottomrule
    \end{tabular}
    }
\end{table}

\begin{table}
    \centering
    \caption{The impact of full fine-tuning on the loss of generalization capabilities. All models are iBOTs pretrained on MillionAID.}
    \label{tab:results-ft-augm}
    \resizebox{0.5\textwidth}{!}{%
    \begin{tabular}{lccccc}
        \toprule
        \textbf{LEVIR-CD: full fine-tuning} & 1:1 & 1:2 & 1:4 & 1:8 & AUC-F1 \\
        \midrule
        iBOT-MillionAID & $88.7 \pm 0.1$ & $86.5 \pm 0.2$ & $63.6 \pm 3.3$ & $7.5 \pm 0.5$ & $67.5 \pm 0.7$ \\
        iBOT-MillionAID-augm & $90.6 \pm 0.2$ & $87.6 \pm 0.9$ & $50.4 \pm 15.1$ & $2.0 \pm 1.0$ & $65.2 \pm 3.2$ \\\midrule
        \textbf{LEVIR-CD: frozen backbone} & & & & & \\
        \midrule
        iBOT-MillionAID & $81.5 \pm 0.1$ & $81.0 \pm 0.4$ & $69.3 \pm 3.1$ & $17.0 \pm 7.9$ & $65.9 \pm 1.6$ \\
        iBOT-MillionAID-augm & $84.4 \pm 0.0$ & $84.4 \pm 0.2$ & $61.6 \pm 7.8$ & $3.4 \pm 4.0$ & $64.7 \pm 2.0$ \\
        \midrule
        \textbf{RESISC45: full fine-tuning} & \multicolumn{4}{c}{} & AUC-ACC \\
        \midrule
        iBOT-MillionAID & $94.6 \pm 0.2$ & $92.8 \pm 0.3$ & $70.4 \pm 4.0$ & $16.6 \pm 4.0$ & $73.7 \pm 1.3$\\
        iBOT-MillionAID-augm &  $93.4 \pm 0.2$ & $84.3 \pm 1.2$ & $47.4 \pm 5.6$ & $18.7 \pm 2.0$ & $66.2 \pm 1.8$ \\
        \midrule
        \textbf{RESISC45: linear probing} & \multicolumn{4}{c}{} &  \\
        \midrule
        iBOT-MillionAID & $91.0 \pm 0.1$ & $87.5 \pm 0.1$ & $60.8 \pm 0.2$ & $9.3 \pm 0.2$ & $68.1 \pm 0.1$ \\
        iBOT-MillionAID-augm &  $94.6 \pm 0.1$ & $92.2 \pm 0.2$ & $66.5 \pm 1.5$ & $25.1 \pm 1.3$ & $73.8 \pm 0.5$ \\
        \midrule
        \textbf{UC Merced: full fine-tuning} & \multicolumn{4}{c}{} &  \\
        \midrule
        iBOT-MillionAID & $98.0 \pm 0.3$ & $97.2 \pm 0.6$ & $87.2 \pm 1.9$ & $38.7 \pm 3.0$ & $82.2 \pm 0.7$ \\
        iBOT-MillionAID-augm & $98.7 \pm 0.8$ & $97.9 \pm 1.3$ & $84.3 \pm 4.3$ & $46.0 \pm 8.3$ & $82.9 \pm 1.0$ \\
        \midrule
        \textbf{UC Merced: linear probing} & \multicolumn{4}{c}{} &  \\
        \midrule
        iBOT-MillionAID & $96.9 \pm 0.0$ & $97.1 \pm 0.2$ & $93.6 \pm 0.2$ & $34.0 \pm 1.3$ & $82.5 \pm 0.2$ \\
        iBOT-MillionAID-augm & $99.5 \pm 0.1$ & $99.2 \pm 0.3$2 & $75.7 \pm 2.9$ & $31.3 \pm 3.9$ & $80.2 \pm 0.7$ \\
        \bottomrule
    \end{tabular}
    }
\end{table}

The results are shown in Figure~\ref{fig:results-auc} and in Table~\ref{tab:results-full}. The general conclusion is that all tested models struggle with generalizability across scales, and none of the methods wins all tasks. General-purpose models like iBOT-ImageNet generally outerperform specialized models on classification tasks and stay a little behind on change detection tasks. 

For the LEVIR-CD dataset, the results are generally comparable across methods. However, GFM shows a clear advantage over the other methods for the 1:2 and 1:4 scale distortions. Specifically, while all four methods produce comparable results at 1:2, GFM demonstrates a clear advantage at 1:4. However, we remark that the pretraining dataset for GFM GeoPile contains RESISC45, which could possibly cause its superior performance over the other methods.
For CDD dataset, we observe that all the results are comparable, however, we observe that GFM does not have superior performance over the other methods. 
The little AUC-F1 score difference between various scale distortions could be explained by the fact that the CDD dataset contains samples from different GSD (0.03m-1m).
For classification, we compare iBOT trained on ImageNet, our trained iBOT for MillionAID, the two versions of Satlas and GFM.
We observe that for iBOT (both trained on ImageNET and MillionAID) linear probing has a clear advantage over full-finetuning for lower resolutions.


In Table~\ref{tab:results-ft-augm}, we report the performance of our trained iBOT on the MillionAID dataset, comparing results with and without augmentations, as well as between a frozen backbone or linear probing and full fine-tuning. 
For change detection on the LEVIR-CD dataset, we observe that full fine-tuning has a clear advantage over a frozen backbone. 
Additionally, we note that augmentations do not improve performance for this task.
For the classification task, 
we observe that for both full fine-tuning and linear probing the model trained with augmentations has a clear advantage over the one trained without augmentation. 

Experiments with augmentations and the results of the default setup for RESISC45 and CDD datasets show that the diversity of the dataset in terms of real resolutions (GSD) improves the generalization capabilities of the finetuned model, even if the backbone weights are frozen.








\ifCLASSOPTIONcaptionsoff
  \newpage
\fi

\bibliographystyle{IEEEtran}
\bibliography{IEEE_GRSL}

\end{document}